%% file: lyl_eemc.tex
\documentclass[10pt, a4paper, twocolumn]{article} 

\input{structure.tex} 

\title{EEMC:Embedding Enhanced Multi-tag Classification} 

\author{
	\authorstyle{Yanlin Li\textsuperscript{1,2} and An shi\textsuperscript{1} and Ruisheng Zhang\textsuperscript{2} } 
	\newline\newline 
	\textsuperscript{1}\institution{Institute of model physics CAS, China}\\ 
	\textsuperscript{2}\institution{Lanzhou University, Gansu, China} \\
}

\date{} 

\begin{document}

\maketitle 

\thispagestyle{firstpage} 


\lettrineabstract{The recently occured representation learning make an attractive performance in NLP and complex network, it is 
becoming a fundamental technology in machine learning and data mining. How to use representation learning to 
improve the performance of classifiers is a very significance research direction. We using representation learning 
technology to map raw data(node of graph) to a low-dimensional feature space. In this space,each raw data obtained 
a lower dimensional vector representation, we do some simple linear operations for those vectors to produce some virtual data,
using those vectors and virtual data to training multitag classifier. After that we meatured the performance of classifier by
F1 score(Macro\% F1 and Micro\% F1). Our method make Macro F1 rise from 28 \% - 450\% and make avarge F1 score rise from
12 \% - 224\%. By contrast, we trained the classifier directly with the lower dimensional vector,and 
meatured the performance of classifiers. We validate our algorithm on three public data sets,we found that the virtual data helped 
the classifier greatly improve the F1 score. Therefore, our algorithm is a effective way to improve the performance 
of classifier. These result suggest that the virtual data generated by simple linear operation, in representation space, 
still retains the information of the raw data. It's also have great significance to the learning of small sample data sets.}


\section{Introduction}
learning also called embedding. Such as "word embeding","graph embeding".
We can think of representation learning as a mapping technology. Many people using data, with the form of vector, to fit their model. 
Obviously vectors is the main "language" in the Machine Learning world. But real world's data "language" are multitudinous. 
How to translate the real world's data to a vector is a primary task for representation learning. 
A good "translate" should be include the underlying structure information. Rl is generated from the actual demand
on one hand, almost whole online data is structured, it's hard to learning knowledge economically and 
fit machine learning model conveniently. People need to find dense and low dimensional vectors to represents graph. 
On the other hand,the embeded graph have many useful properties: distance for embeded feature space own latent meaning. 
Researchers found the distance between "China" and "Beijing" equal to the distance between "Japan" and "Tokoy"\cite{EEMC8}. 
Words with the same label, will occur at the same region in the embedded feature space. 

In 2013, R. Al-Rfou \cite{EEMC8} proposed word2vec model, that is a extended skipgram model which mapped words to some vectors.
In the vector space, words have the same semantics have to located on same region. Even the distance between two different class word vectors 
has a specific meaning. All this shows a powerful potention for representation learning. In word2vec, the relationalship 
between words be considered as a conditional probability, the trainning objective of word2vec is to maximum this probability, 
\begin{equation}\frac{1}{T}\sum_{t=1}^T \sum_{-c<=j<=c,j\ne 0} \log P(\omega_{t+j}|\omega_{t} ) \label{E1}\end{equation}

where $c$ is the size of the training contex. $\omega_{1}$, $\omega_{2}$, $\omega_{3}$?\dots $\omega_{T}$ are a sequence of training words. $\omega_{t}$ is called centre word.
It means a word have been known, $\omega_{t+j}$ are called surrounding words. they represents the context with centre word.

In another area of research, analysis graph data is a import work for bigdata application. 
graph is structured data, usually using adjacent matrix to represented the relationalship for nodes.
With the continuous grows in graph size, using adjacent matrix to represents graph data becomeing more an more unrealistic, 
since when $n$ nodes are added, the size of adjacent matric will increase $n^{2}$.
Deepwalk\cite{EEMC9} inspired by word2vec, this algorithm consider nodes as words, and take a random walk on graph to sample a sequence
of nodes, the sequence can be treated like a sentence. In this model, the optimization problem is similarity as formula 1, 
and training process is almost same as word2vec.
\begin{equation} \min_{\Phi} \quad  -log P(v_{i-\omega},\dots,v_{i+\omega} \setminus v_{i} \quad| \Phi_{v_{i}} ) \label{E2}\end{equation}
where,$\Phi_{v_{i}}$ is the mapping of the node $v_{i}$, that is a embeded node,$v_{i}$ like a centre word in a training windows. $\omega$
is the width of training windows. This discovery leads to the wider use of structured data in machine learning.

Mutilabel classification method are fundamental required in graph embedding expriment, it also have great uses in structured 
data classification, such as social network analysis, protein function classification, and intelligent recommendation and 
photo classification. Grigorios Tsoumakas et al, consider that muti-label classification methods can be divided into two main categories:
\begin{itemize}
	\item [1)]Problem transformation methods. Problem transformation methods transform the multi-label classification problem 
  either into one or more single-label classification or regression problems. In this way, single-label classifiers are employed; 
  and their singlelabel predictions are transformed into multi-label predictions. Problem transformation is attractive on account 
  of both scalability and flexibility: any off-the-shelf single-label classifier can be used to suit requirements. 
	\item [2)]Algorithm adaptation methods. The algorithm adaptation methods, that extend specific learning algorithms in order to 
	handle multi-label data directly \cite{EEMC1}. Well-known approaches include AdaBoost, decision trees. Such
	methods are usually chosen to work specifically in certain domains.
\end{itemize}

\begin{figure}
  \includegraphics[width=1.1\linewidth]{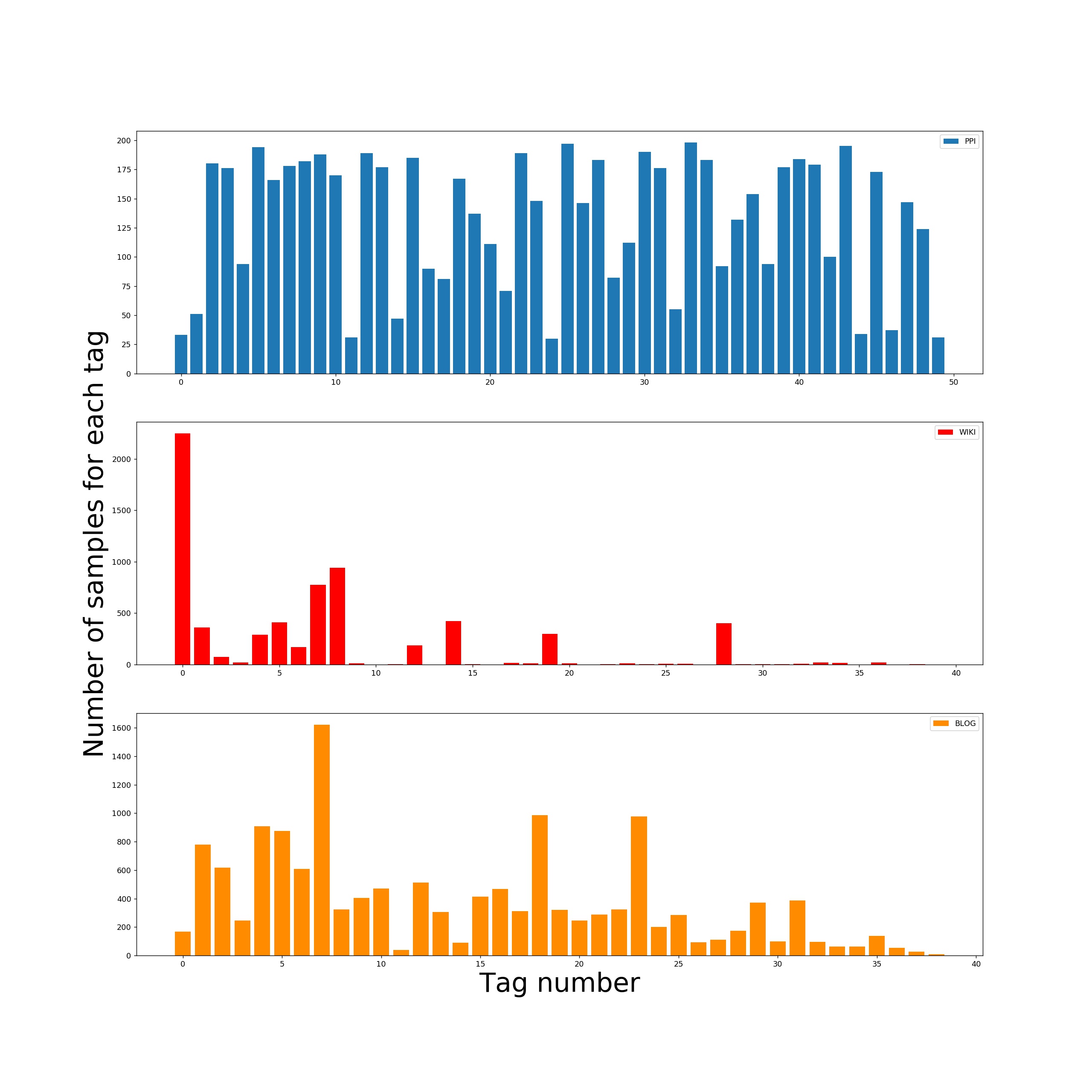}  
	\centering
  \caption{Three graph dataset are imbalanced} 
  \label{imbalance} 
\end{figure}

Multi-tag classification tasks are commonly used methods for evaluating embedding quality. 
This method is usually trained in semi-supervised mode.
Semi-supervised means that only part of the data is labeled. In actual data, part of the data 
is usually selected as labeled data, and the remaining data labels are predicted by the trained model.
Actualy, under-fitting or non-working of a certain sub-classifier can easily occur, when the number of 
samples is not large. This situation actually affects the evaluation of embedded quality,Even the best classifier, 
the output result in the absence of data is disastrous. 
The imbalance of the data set is the macro cause of this phenomenon. As show in figure \ref{imbalance}, 
The number of samples belonging to one label in the data set is likely to be hundreds or even thousands of times that of another.
A simple idea to deal with the proplem is to copy the data directly, so that the label classes with a small sample 
will meet the training requirements. Actually the result were not ideal, this is because simply copying data does not increase 
the diversity of data in a label class. As we'll see later in the experiment, the benefits of simply adding data are minimal. 

\begin{figure}
	\includegraphics[width=0.6\linewidth]{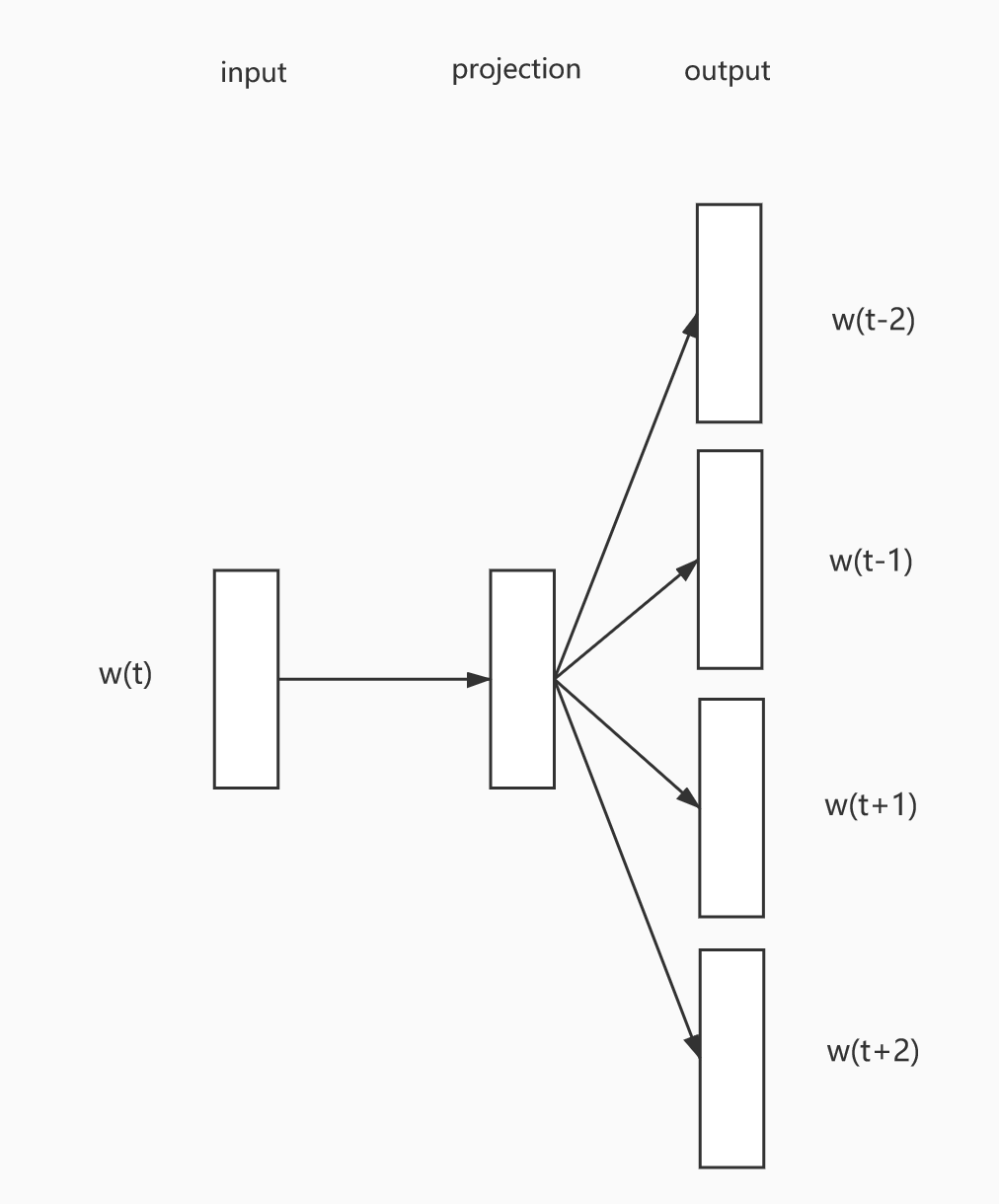}  
	\centering
	\caption{skipgram model architecture} 
	\label{w2v} 
\end{figure}

\section{Algorithm Framework}
\subsection{Embedding Node to Feature Space}
General, the random walk category model reference the SkipGram of Word2vec which is the classic NLP algorithm.
We take the node as a word, take the sampling node sequence as sentence. 
We fomulate the model as a maximum likelihood optimization problem just like a langure model. 
Let $G=(V,E)$ be a given network. Let $f:V \to R^d $ be the mapping functionn from node to feature space.
$d$ is a parameter specifying the number of dimensions of the feature space. 
Actually, $f$ is an $\left| V \right| \times d$ matrix. Our goal is to learn a vectors to every node, 
we using an auto-encoder to learn the topological of the networks from the "sentence" of the nodes, then seek to optimize the following objective function:

\begin{equation}\max_{f} \quad \sum_{v\in V}\log P(N_{i}(v)| f(v) ) \label{E3}\end{equation} Where $N_{i}(v)$ is the 
neighbor nodes set of source node $v$. We assume that the neighorhood of the source node are indepandence 
for each other. Then we can simplify the likelihood probability function:
\begin{equation}P(N_{i}(v)|f(v)) = \prod_{n_{i}\in N_{i}(v )} P(n_{i}| f(v) ) \label{E4}\end{equation} 
	
We using a sigmod function to map the dot product of the mapped vectors to a  probability between 0 and 1:
\begin{equation}P(n_{i}| f(v) )=\frac{1}{(1+\exp(f(v)\cdot f(n_{i}))        )}  \label{E5}\end{equation} 
The equ.\ref{E5} represents the similarities between source node and neighorhood node.  

\subsection{Autocoder Model}
Generally, autocoders can be used for dimensionality reduction. It's consists of encoder and decoder. 
The autocoder can be described as:
\begin{equation}g(f(\vec{x}))=\vec{x}\label{E6}\end{equation}
The training process can be described as:
\begin{equation}min \quad L(\vec{x},g(f(\vec{x}))\label{E7}\end{equation}
$L$ is a loss function, used to indicate the gap between $\vec{x}$ and $g(f(\vec{x})$. 
The training process is minimzing $L$ function.

\subsection{Multitag Classification And Virtual Data}

Multitag classification is the basic task for bigdata application. 
Node classification is a benchmark to evaluate the embedding quality in the field of complex network and network embedding.
A large number of scholars only focus on the quality of embedding, but ignore the impact of the data itself on the performance 
of the classifier. In fact, it is difficult for classifiers to work properly when the data itself is seriously unbalanced.
for example, the dataset only have one sample is labeled "A", at the same time, hundreds of samples are labeled as "B".
so the classifier in charge of data labeled "A" will work badly, since we have no enough data to fit it.
This phenomenon is very common in small and medium-sized networks.

Based on the observation of previous work, we notice that the embedded data has nice properties. 
For instance, the nodes have same label will embeded more closer in feature space. Meanwhile, 
distance represents the degree of similarity between samples. In other words, the distance between two samples with 
same label far less than the sample with different label. The same labeled samples have almost same distance far from origin. 
Base on these phenomena, we make the following hypothesis:
\begin{itemize}
	\item Same label samples will be embedded in a compacted space. 
	\begin{equation} \label{E8}
		D(\vec{x_{L}},\vec{y_{L}}) \ll D(\vec{x_{L}},\vec{y_{L^{'}}})
	\end{equation}
	\item Distance is the only measure of samples' similarity. 
	\begin{align}\label{E9}
 \vec{x}\in L,\quad  \vec{y} \in L \Leftrightarrow  D(\vec{x},\vec{0}) \approx D(\vec{y},\vec{0})
	\end{align}
	
	\item The distance  between  different label samples much bigger than same labe samples. 
\end{itemize}
    \begin{equation}\label{E10}
		|D(\vec{x_{L}},\vec{0})-D(\vec{y_{L^{'}}},\vec{0})|\gg 0 
	\end{equation}
	
We know the embedding model only have one single hidden layer, this can be express as fellow:
\begin{equation}g(\vec{x})=f(\omega*\vec{x}+b)) \label{E11}\end{equation}
$g(\vec{x})$ is a function in embedded space. $ f(\vec{x})$ is sigmod function.

\begin{flalign*}
	\begin{split}
		&\quad g(\theta\vec{x_{1}}+(1-\theta)\vec{x_{2}})\\\\
		&=f(\omega(\theta\vec{x_{1}}+(1-\theta)\vec{x_{2}})+b)\\\\
		&=f(\theta(\omega\vec{x_{1}}+b)+(1-\theta)(\omega\vec{x_{2}}+b))\\
	\end{split}&
\end{flalign*}
Since $f(\vec{x})$ is convex,  when $\vec{x} < 0$, 

\begin{flalign*}
	\begin{split}
		&\quad f(\theta(\omega\vec{x_{1}}+b)+(1-\theta)(\omega\vec{x_{2}}+b))\\\\
		&\ \leq \theta f(\omega\vec{x_{1}}+b)+(1-\theta)f(\omega\vec{x_{2}}+b)\\\\
		&\ =\theta g(\vec{x_{1}})+(1-\theta)g(\vec{x_{2}})\\
	\end{split}&
\end{flalign*}
So,  when $\vec{x} < 0$,
\begin{equation}
	g(\theta\vec{x_{1}}+(1-\theta)\vec{x_{2}}) \leq \theta g(\vec{x_{1}})+(1-\theta)g(\vec{x_{2}})\label{E12}
\end{equation}
By the same token, when $\vec{x} > 0$,
\begin{equation}
	g(\theta\vec{x_{1}}+(1-\theta)\vec{x_{2}}) \geq \theta g(\vec{x_{1}})+(1-\theta)g(\vec{x_{2}})\label{E13}
\end{equation}

\begin{figure}
	\includegraphics[width=1\linewidth]{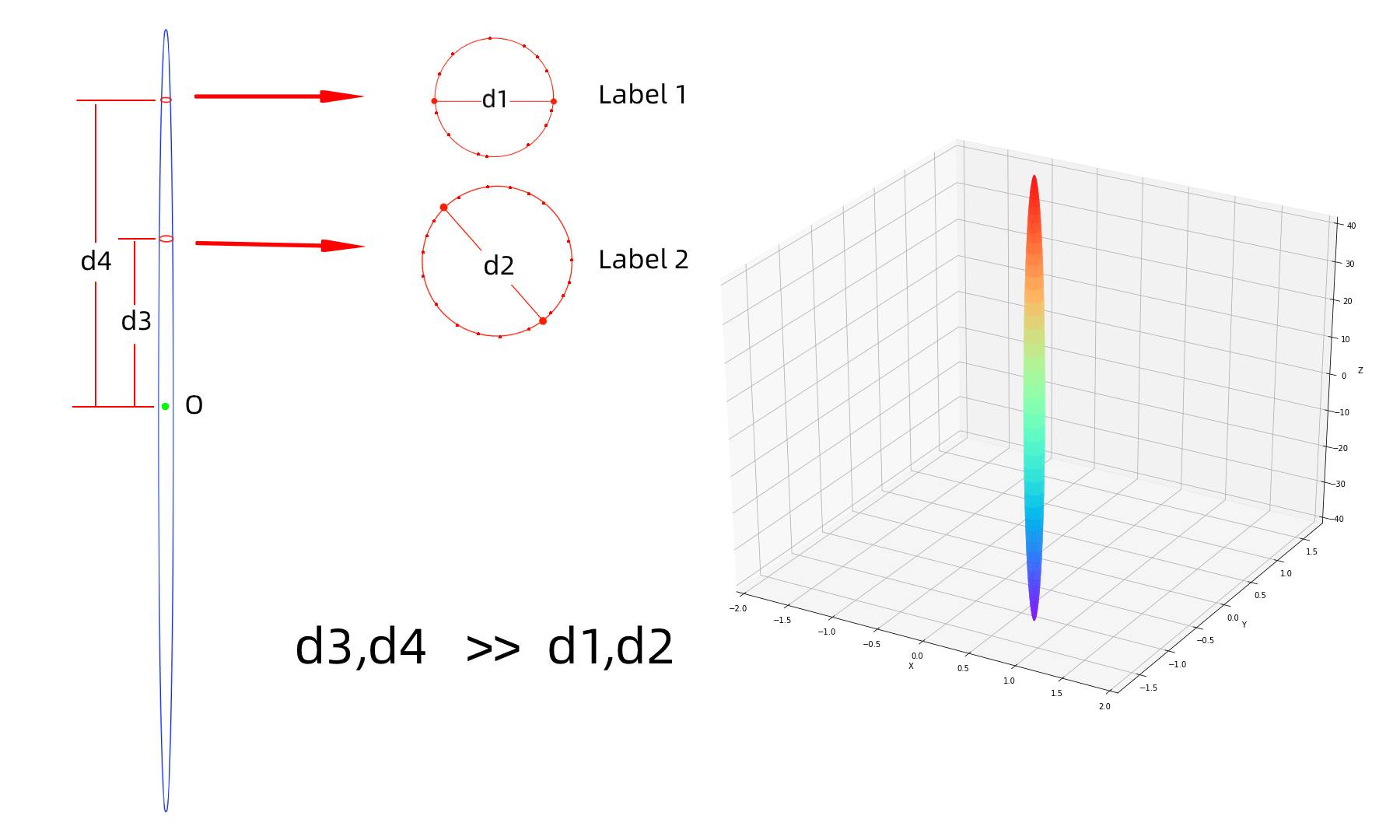}  
	\centering
	\caption{3D Feature space} 
	\label{space} 
\end{figure}

Therefore, $g(\vec{x})$ is convex, when $\vec{x} < 0$ and is concave, when $\vec{x} > 0$.
We assume that $g(\vec{x})$ is in three dimensional space. 
So $g(\vec{x})$ like a Ellipsoid surface with very short a, b axis and very long c axis show in Figure\ref{space}. 
This inference is based on equ. \ref{E8} and equ. \ref{E10} and just calculated result.

Assume that $\vec{v_{1}}$ and $\vec{v_{2}}$ located on a ring of the surface, because they are same labeled 
so,
\begin{equation}\label{E14}
	D(\vec{v_{1}},\vec{0})\approx D(\vec{v_{2}},\vec{0})
\end{equation}
We using this two samples to produce a virtual data $\vec{v_{3}}$
\begin{equation}\label{E15} \vec{v_{3}} = \theta \vec{v_{1}} + (1-\theta) \vec{v_{2}}  \end{equation}
Since equ.\ref{E14} and equ.\ref{E8} using triangle rule we can prove
\begin{equation}\label{E16}
	D(\vec{v_{3}},\vec{0})\approx D(\vec{v_{1}},\vec{0})\approx D(\vec{v_{2}},\vec{0})
\end{equation}
because of equ. \ref{E9} , we can infer, 
\begin{equation}
	\vec{v_{3}} \in L 
\end{equation}

This properties imply us that we can product any number of data using embedded nodes. 
However, in order to maintaining the diversity of samples in the same label set we select $\theta =0.5$,
so we produce a virtual node data in feature space like:
\begin{equation}\quad V_{node}=\frac{\vec{v_{1}}+\vec{v_{2}}}{2}  \label{E17}\end{equation}  
we randomly choice two existing samples, then using equ.\ref{E17} to generate new data. 
Repeat the process, until the number of data in every tag set is equal to $addcoeff$ * $num_{max}$. 
The value of $num_{max}$ present the maxium number of data in all tag dataset, $addcoeff$ is a coefficient between 0 and 1.

  \section{Experiment}
  In this section we present an experimental analysis of our method. 
  we test our algorithm in PPI, WiKi, and Blogcatalog graph to classify nodes. 
  In this graph, every node get one or more labels from a label set $L$. We using semi-supervised
  method to training the classifier. All expriments started from the number of 400 nodes and gradually 
  increased 200 nodes per step. We set the parameter: $size=120 $,$windows=5$,read sampled files, in this 
  file store sampled sequence by random walk.
  
  \subsection{PPI Classification}
\begin{figure*}
    \includegraphics[width=1\linewidth]{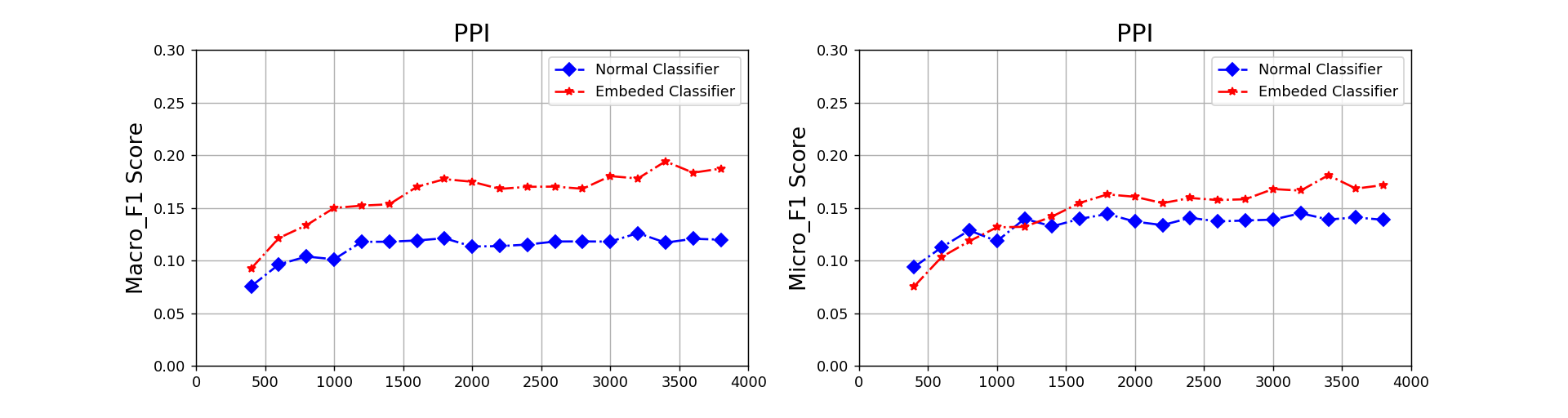}  
    \centering
    \caption{PPI classify result1} 
    \label{PPI} 
  \end{figure*}

  Protein-Protein Interactions: PPI is a subgraph of Homo Sapiens. The subgraph corresponds to the graph 
  induced by nodes for which we could obtain labels from the hallmark gene sets and represent biological states.
  The PPI graph has 3890 nodes and 76584 edges and 50 class labels. In our test, we trained data start from 400 nodes 
  to 3800 nodes, every step increase 200 nodes. We set the parameter: $addcoeff$=1 to add the virtual nodes and set $nodenum=3890$. 
  As show in figure \ref{PPI}, comparing with normal method(trained with no virtual nodes), our method make a significant 
  improvement in F1 score, it achieve a gain of 43.9\% in Macro\_F1 score and 11.17\% in Micro\_F1 score over normal method.  
  
  \subsection{WIKI Classification}
\begin{figure*}
    \includegraphics[width=1\linewidth]{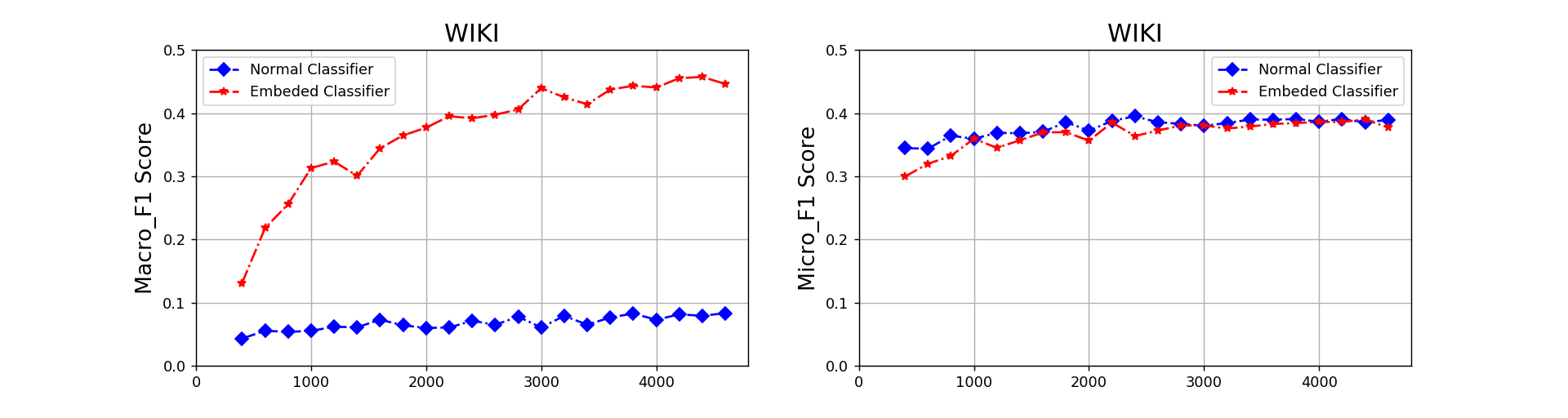}  
    \centering
    \caption{Wikipedia classify result1} 
    \label{wiki} 
  \end{figure*}

  WIKI is a cooccurrence network of words appearing in the first million bytes of the Wikipedia dump.
  The labels represent the Part-of-Speech (POS) tags inferred using the Stanford POS-Tagger.
  The graph has 4,777 nodes, 184,812 edges, and 40 different labels.
  we set the parameter: $addcoeff$=0.35 to add the virtual nodes and set $nodenum=4777$. 
  As show in figure \ref{wiki}, comparing with normal method(trained with no virtual nodes),
  our algorithm achieve a tremendous gain of 452\% in Macro\_F1 score even with a slight reduction
  3.16\% in Micro\_F1 score. To sum up, our method giving us 224\% gain over normal method in avarge F1 score.     
  
  \subsection{BLOG Classification}
\begin{figure*}
    \includegraphics[width=1\linewidth]{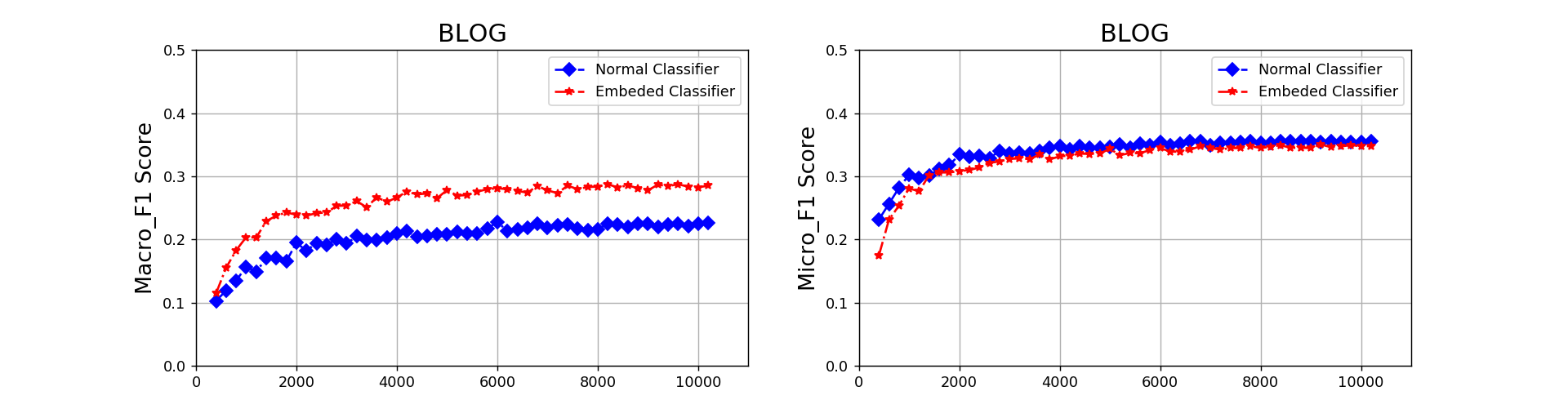}  
    \centering
    \caption{Blogcatalog classify result1} 
    \label{blog} 
  \end{figure*}

  This is a graph of social relationships of the bloggers showed on the BlogCatalog website. The labels
  represent blogger interests inferred through the metadata provided by the bloggers. 
  The network has 10,312 nodes,333,983 edges,and 39 different labels.
  We set the parameter: $addcoeff$=0.15 to add the virtual nodes and set $nodenum=10312$. 
  As show in figure \ref{blog}, comparing with normal method(trained with no virtual nodes),
  our algorithm achieve a distinct gain of 28.7\% in Macro\_F1 score even with a slight reduction
  3.56\% in Micro\_F1 score. To sum up, our method giving us 12.5\% gain over normal method in avarge F1 score.
  
  \subsection{Experiment Analyzing}
  In our expriments, an obvious rule is, the small for data, the better for result. 
  This situation is also understandable, a smaller dataset have a high probability of getting underfitting subclassifiers.      
  Conversely, a big dataset have low probability of getting underfitting subclassifiers. 
  Expriment result shows that, our method is effective in improving the accuracy of classifier.  
  In our expriment, we set different value to $addcoeff$ to achieve the best result.
  This parameter determines the size of virtual nodes need to be add in training dataset. 
  A certain fact will be mentioned, this parameter is not proportional to F1 score. An reasonable value
  will got higher score.  
  
  \subsection{Parameter Sensitivity}
\begin{figure*}
    \includegraphics[width=1 \linewidth] {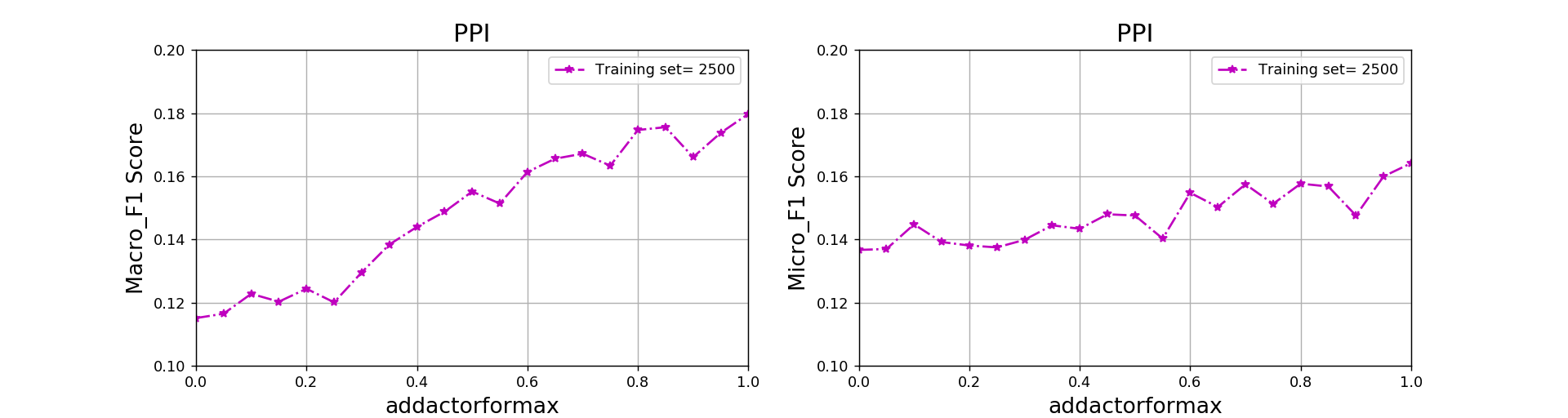}  
    \centering
    \caption{Impact of virtual data number in PPI} 
    \label{pm_ppi} 
  \end{figure*}

\begin{figure*}
    \includegraphics[width=1 \linewidth] {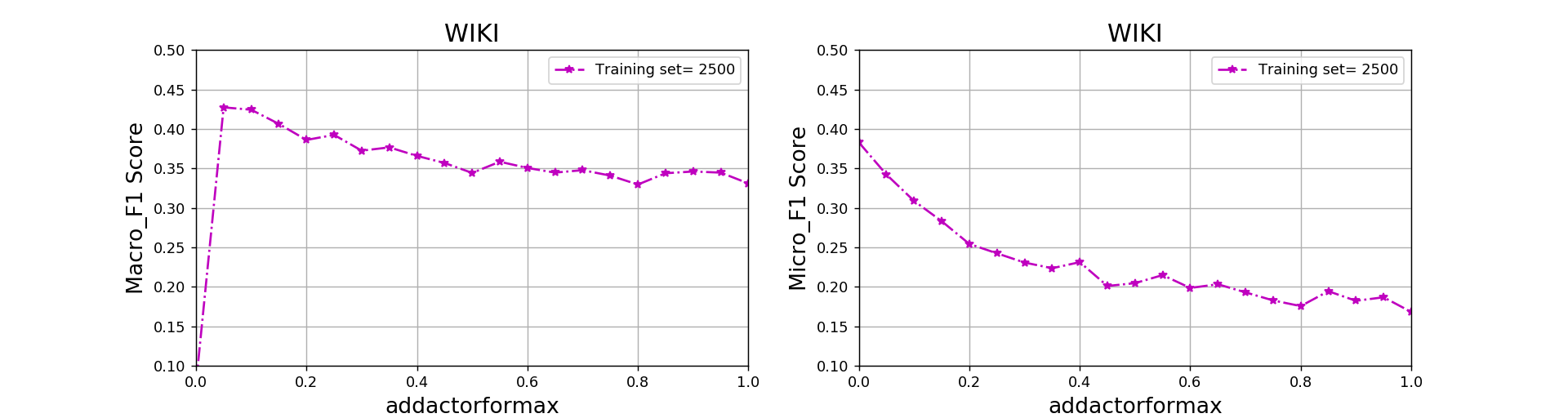}  
    \centering
    \caption{Impact of virtual data number in WIKI} 
    \label{pm_wiki} 
  \end{figure*}

\begin{figure*}
    \includegraphics[width=1 \linewidth] {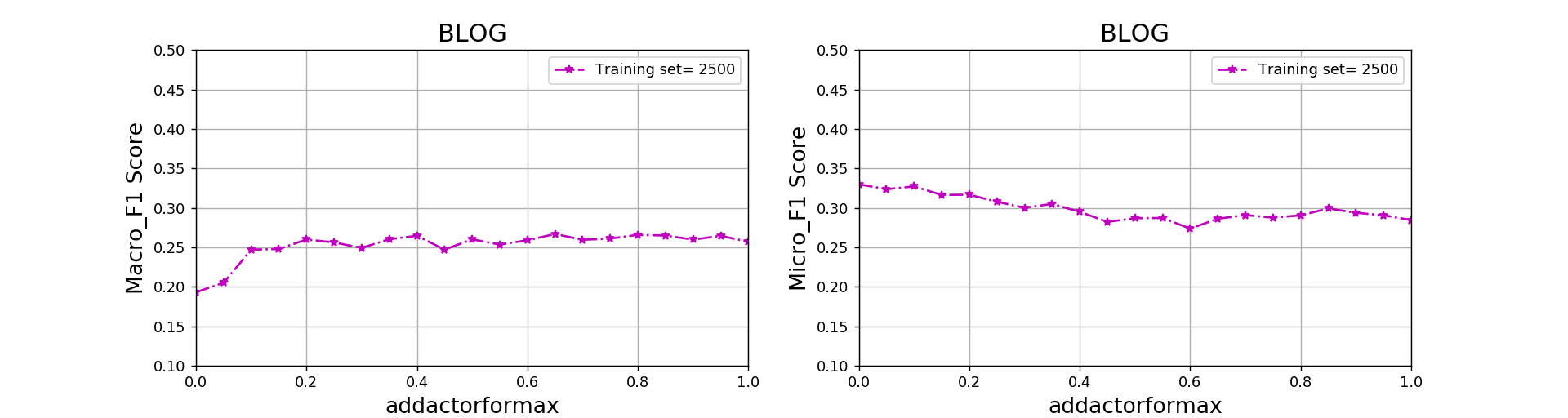}  
    \centering
    \caption{Impact of virtual data number in BLOG} 
    \label{pm_blog} 
  \end{figure*}

  In order to evaluate the impact of parameters for classification performance of EEMC.
  We design experiments on three multi-label classification tasks. Actually the parameter $addcoeff$ is a factor 
  from 0 to 1. Since, only the $addcoeff$ directly related to the fixed dataset, so we fixed the embeded 
  parameter(window size and the walk length) and training number of nodes ($train_num$=1200).
  We then vary the size of factor $addcoeff$ to obvering it's impact on the classification performance. 
  As show in Figure\ref{pm_ppi}, With the increase of virtual data, macro F1 and micro F1 scores are all increasing.
  In Figure\ref{pm_wiki}, the macro F1 score first soared sharply,then gradually decrease with the parameter $addcoeff$ 
  increases. The micro F1 score gradually decrease with the parameter $addcoeff$ increases, this suggests that it is important 
  to select an appropriate parameter on some dataset. 
  In Figure\ref{pm_blog}, the macro F1 score and micro F1 score are also change slow, macro F1 first rise and then leveled off,
  with the parameter $addcoeff$ increases. micro F1 score make a slow decline then leveling off.
  Although the micro F1 in Figure\ref{pm_wiki} and Figure\ref{pm_blog} are all slow decline, but the accuracy of the 
  classification can still benefit from macro F1 and avarge F1.
  Such as in wiki expriment, when we set $addcoeff=0.35$, the micro F1 score decline 3\% but macro F1 score gain of 452\%.   
  This experiment also suggests that our approach works best for small and middle datasets with more class tags.  

  Refer to figure \ref{imbalance}, in PPI dataset, the label class with the most data has about 
  200 samples. In Wiki and Blog this number are about 2400 and 1600. In PPI we set $addcoeff =1$.
  in WIKI $addcoeff=0.35$, in Blog $addcoeff=0.15$. Combined with experimental results, we can infer that the number of 
  trainning samples less than 600 and more than 200 is appropriate for classifiers. More than this number will cause overfitting.
  In Blog dataset, most label class have enough samples to fit classifiers, so the benefits from our algorithm are minimal.

  
\section{Conclusion}
In this paper, we propose a general method to boost the performance for off-the-shelf classifier,
our method try to generate some virtual data for those labels that include few sample. Use this data
to training classifier,then we have proved the validity of our method through three expriments,
experimental results show that our method is remarkably effective in improving the performance of classifier.
We also test the sensitivity of the parameters, the experimental results shows that on partial dataset,
bigger parameter is not always better, set an appropriate parameter is very import. Since the embedding process
does not require a large sample, and after embedding, meaningful virtual data can be easily obtained. So our method 
also have great significance to the learning of small sample data sets. In conclusion, our method can significantly 
improve the performance of the classifier, meanwhile, our method explored another way of how improving the performance of classifier.




\end{document}

%% file: structure.tex
\usepackage[english]{babel} 

\usepackage{microtype} 

\usepackage{amsmath,amsfonts,amsthm} 

\usepackage[svgnames]{xcolor} 

\usepackage[hang, small, labelfont=bf, up, textfont=it]{caption} 

\usepackage{booktabs} 

\usepackage{lastpage} 

\usepackage{graphicx} 

\usepackage{enumitem} 
\setlist{noitemsep} 

\usepackage{sectsty} 
\allsectionsfont{\usefont{OT1}{phv}{b}{n}} 

\usepackage{geometry} 

\usepackage[T1]{fontenc} 
\usepackage[utf8]{inputenc} 

\usepackage{XCharter} 

\usepackage{fancyhdr} 
\fancypagestyle{firstpage}{ 
	\fancyhf{}
}

\newcommand{\authorstyle}[1]{{\large\usefont{OT1}{phv}{b}{n}\color{DarkRed}#1}} 

\newcommand{\institution}[1]{{\footnotesize\usefont{OT1}{phv}{m}{sl}\color{Black}#1}} 

\usepackage{titling} 

\newcommand{\HorRule}{\color{DarkGoldenrod}\rule{\linewidth}{1pt}} 

\pretitle{
	\vspace{-30pt} 
	\HorRule\vspace{10pt} 
	\fontsize{32}{36}\usefont{OT1}{phv}{b}{n}\selectfont 
	\color{DarkRed} 
}

\posttitle{\par\vskip 15pt} 

\preauthor{} 

\postauthor{ 
	\vspace{10pt} 
	\par\HorRule 
	\vspace{20pt} 
}

\usepackage{lettrine} 
\usepackage{fix-cm}	

\newcommand{\initial}[1]{ 
	\lettrine[lines=3,findent=4pt,nindent=0pt]{
		\color{DarkGoldenrod}
		{#1}
	}{}%
}

\usepackage{xstring} 
\newcommand{\lettrineabstract}[1]{
	\StrLeft{#1}{1}[\firstletter] 
	\initial{\firstletter}\textbf{\StrGobbleLeft{#1}{1}} 
}

\usepackage[backend=bibtex,natbib=true]{biblatex} 

\usepackage[autostyle=true]{csquotes} 